\documentclass[runningheads]{llncs}
\usepackage{graphicx}

\usepackage{tikz}
\usepackage{comment}
\usepackage{amsmath,amssymb} 
\usepackage{color}
\usepackage{hyperref}
\hypersetup{
  colorlinks, linkcolor=red
}

\usepackage{bm}
\usepackage{booktabs}
\usepackage{cite}

\usepackage{multirow}
\usepackage{wrapfig}
\usepackage{enumerate}
\usepackage{subfigure}
\usepackage{algorithm}
\usepackage{algpseudocode}
\usepackage{bbding}
\usepackage{wrapfig}

\usepackage[accsupp]{axessibility}  

\begin{document}
\pagestyle{headings}
\mainmatter
\def\ECCVSubNumber{1175}  

\title{Label2Label: A Language Modeling Framework for Multi-Attribute Learning} 

\titlerunning{Label2Label}
%
\author{Wanhua Li \and
Zhexuan Cao \and
Jianjiang Feng \and
Jie Zhou \and
Jiwen Lu\thanks{Corresponding author}}
\authorrunning{W. Li et al.}
%
\institute{Department of Automation, Tsinghua University, China \and
Beijing National Research Center for Information Science and Technology, China \\
\email{ \{wanhua016,caozx00\}@gmail.com; \{jfeng,jzhou,lujiwen\}@tsinghua.edu.cn}}
\maketitle

\begin{abstract}
  Objects are usually associated with multiple attributes, and these attributes often exhibit high correlations. Modeling complex relationships between attributes poses a great challenge for multi-attribute learning. This paper proposes a simple yet generic framework named Label2Label to exploit the complex attribute correlations. Label2Label is the first attempt for multi-attribute prediction from the perspective of language modeling.
  Specifically, it treats each attribute label as a ``word'' describing the sample. As each sample is annotated with multiple attribute labels, these ``words'' will naturally form an unordered but meaningful ``sentence'', which depicts the semantic information of the corresponding sample. Inspired by the remarkable success of pre-training language models in NLP, Label2Label introduces an image-conditioned masked language model, which randomly masks some of the ``word'' tokens from the label ``sentence'' and aims to recover them based on the masked ``sentence'' and the context conveyed by image features. Our intuition is that the instance-wise attribute relations are well grasped if the neural net can infer the missing attributes based on the context and the remaining attribute hints. Label2Label is conceptually simple and empirically powerful. Without incorporating task-specific prior knowledge and highly specialized network designs, our approach achieves state-of-the-art results on three different multi-attribute learning tasks, compared to highly customized domain-specific methods. Code is available at \url{https://github.com/Li-Wanhua/Label2Label}.
\end{abstract}

\section{Introduction}
\label{sec:intro}
Attributes are mid-level semantic properties for objects which are shared across categories~\cite{farhadi2009describing,duan2012discovering,li2021learning,feris2017visual}. We can describe objects with a wide variety of attributes. For example, human beings easily perceive gender, hairstyle, expression, and so on from a facial image~\cite{li2019bridgenet,li2020social}.  Multi-attribute learning, which aims to predict the attributes of an object accurately, is essentially a multi-label classification task~\cite{shin2020semi}. As multi-attribute learning involves many important tasks, including facial attribute recognition~\cite{liu2015deep,kalayeh2017improving,cao2018partially}, pedestrian attribute recognition~\cite{jia2021spatial,tang2019improving,guo2019visual}, and cloth attribute prediction~\cite{liu2016deepfashion,zhang2020texture}, it plays a central role in a wide range of applications, such as face identification~\cite{cao2018partially}, scene understanding~\cite{shao2015deeply}, person retrieval~\cite{li2018richly}, and fashion search~\cite{ak2018learning}.

For a given sample, many of its attributes are correlated. For example, if we observe that a person has blond hair and heavy makeup, the probability of that person being attractive is high. Another example is that the attributes of beard and woman are almost impossible to appear on a person at the same time. Modeling complex inter-attribute associations is an important challenge for multi-attribute learning. To address this challenge, most existing approaches~\cite{rudd2016moon,cao2018partially,jia2021spatial,tang2019improving} adopt a multi-task learning framework, which formulates multi-attribute recognition as a multi-label classification task and simultaneously learns multiple binary classifiers. To boost the performance, many methods further incorporate domain-specific prior knowledge.  For example, PS-MCNN~\cite{cao2018partially} divides all attributes into four groups and presents highly customized network architectures to learn shared and group-specific representations for face attributes. In addition, some methods attempt to introduce additional domain-specific guidance~\cite{kalayeh2017improving} or annotations~\cite{liu2016deepfashion}. However, these methods struggle to model sample-wise attribute relationships with a simple multi-task learning framework.

\begin{figure}[t]
  \centering
  \subfigure[Existing Multi-task Learning Framework]{
    \label{fig:motivation:MTL}
    \includegraphics[width=0.97\linewidth]{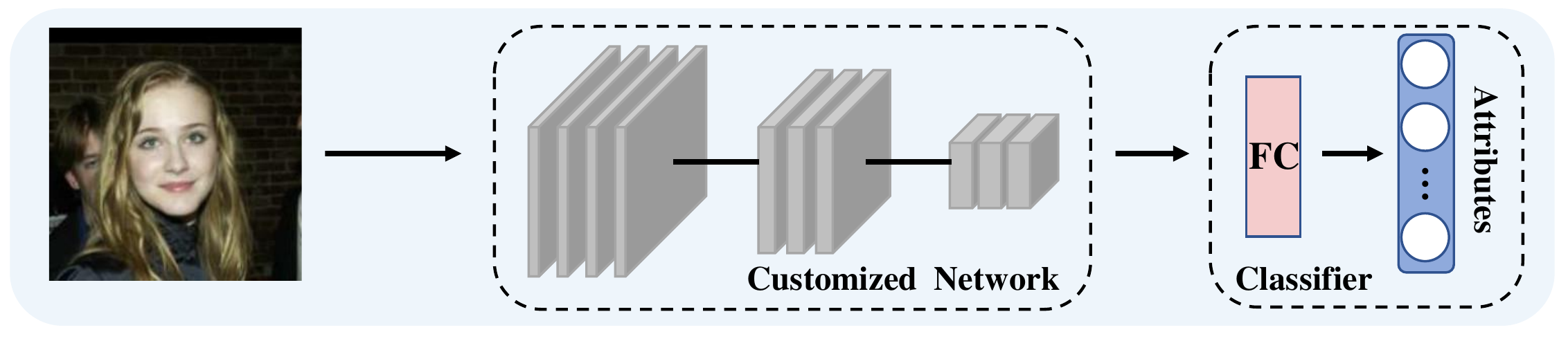}
  }
  \subfigure[Our Language Modeling Framework]{
    \label{fig:motivation:MLM}
    \includegraphics[width=0.97\linewidth]{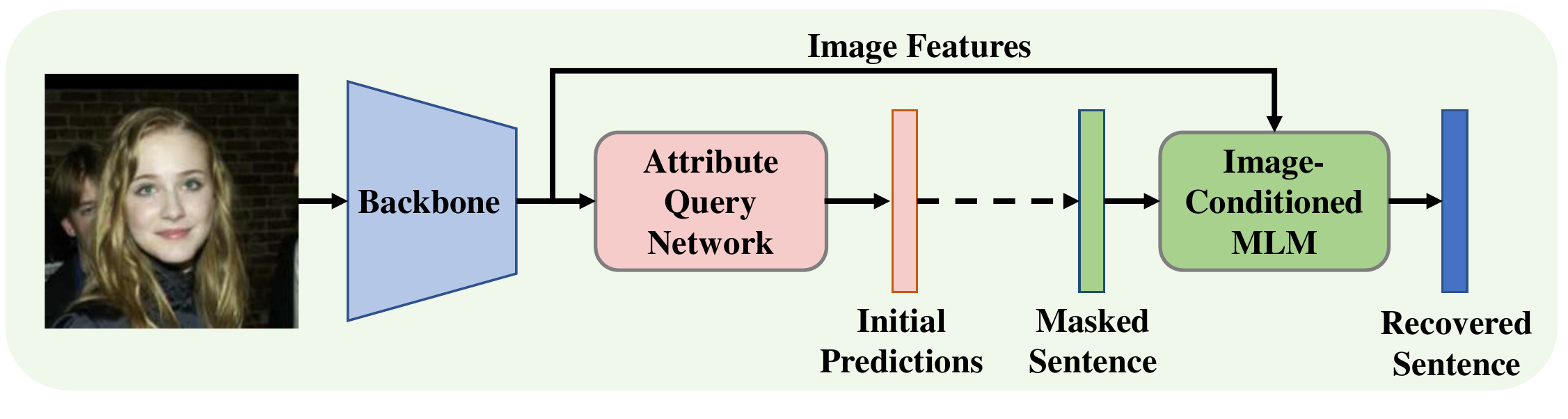}
  }
  \caption{Comparisons of the existing multi-task learning framework and our proposed language modeling framework.
  }
  \label{fig:motivation}
\end{figure}

Recent years have witnessed great progress in the large-scale pre-training language models~\cite{peters2018deep,brown2020language,devlin2018bert}. As a representative work, BERT~\cite{devlin2018bert} utilizes a masked language model (MLM)~\cite{taylor1953cloze} to capture the word co-occurrence and language structure. Inspired by these methods, we propose a language modeling framework named Label2Label to model the complex instance-wise attribute relations. Specifically, we regard an attribute label as a ``word'', which describes the current state of the sample from a certain point of view. For example, we treat the labels ``attractive'' and ``no eyeglasses'' as two ``words'', which give us a sketch of the sample from different perspectives. As multiple attribute labels of each sample are used to depict the same object, these ``words'' can be organized as an unordered yet meaningful ``sentence''. For example, we can describe the human face in Fig. \ref{fig:motivation} with the sentence ``attractive, not bald, brown hair, no eyeglasses, not male, wearing lipstick, ...''. Although this ``sentence'' has no grammatical structure, it can convey some contextual semantic information. By treating multiple attribute labels as a ``sentence'', we exploit the correlation between attributes with a language modeling framework.

Our proposed Label2Label consists of an attribute query network (AQN) and an image-conditioned masked language model (IC-MLM). The attribute query network first generates the initial attribute predictions. Then these predictions are treated as pseudo label ``sentences'' and sent to the IC-MLM. Instead of simply adopting the masked language modeling framework, our IC-MLM randomly masks some ``word'' tokens from the pseudo label ``sentence'' and predicts the masked ``words'' conditioned on the masked ``sentence'' and image features. The proposed image-conditioned masked language model provides partial attribute prompts during the precise mapping from images to attribute categories, thereby facilitating the model to learn complex sample-level attribute correlations.
We take facial attribute recognition as an example and show the key differences between our method and existing methods in Fig. \ref{fig:motivation}.

We summarize the contributions of this paper as follows:
\begin{itemize}
\item
  We propose Label2Label to model the complex attribute relations from the perspective of language modeling. As far as we know, Label2Label is the first language modeling framework for multi-attribute learning.
\item
  Our Label2Label proposes an image-conditioned masked language model to learn complex sample-level attribute correlations, which recovers a ``sentence'' from the masked one conditioned on image features.
\item
  As a simple and generic framework, Label2Label achieves very competitive results across three multi-attribute learning tasks, compared to highly tailored task-specific approaches.
\end{itemize}

\section{Related Work}

\textbf{Multi-Attribute Recognition:} Multi-attribute learning has attracted increasing interest due to its broad applications~\cite{cao2018partially,li2018richly,ak2018learning}. It involves many different visual tasks~\cite{hand2017attributes,jia2021spatial,liu2016deepfashion} according to the object of interest. Many works focus on domain-specific network architectures. Cao \emph{et al.}~\cite{cao2018partially} proposed a partially shared multi-task convolutional neural network (PS-MCNN) for face attribute recognition. The PS-MCNN consists of four task-specific networks and one shared network to learn shared and task-specific representations.
Zhang \emph{et al.}~\cite{zhang2020texture} proposed Two-Stream Networks for clothing classification and attribute recognition.
Since some attributes are located in the local area of the image, many methods~\cite{guo2019visual,sarafianos2018deep,tang2019improving} resort to the attention mechanism. Guo \emph{et al.}~\cite{guo2019visual} presented a two-branch network and constrained the consistency between two attention heatmaps. A multi-scale visual attention and aggregation method was introduced in~\cite{sarafianos2018deep}, which extracted visual attention masks with only attribute-level supervision. Tang \emph{et al.}~\cite{tang2019improving} proposed a flexible attribute localization module to learn attribute-specific regional features. Some other methods~\cite{kalayeh2017improving,liu2016deepfashion} further attempt to use additional domain-specific guidance. Semantic segmentation was employed in~\cite{kalayeh2017improving} to guide the attention of the attribute prediction. 
Liu \emph{et al.}~\cite{liu2016deepfashion} learned clothing attributes with additional landmark labels. There are also some methods~\cite{zhao2019recognizing,shu2021learning} to study multi-attribute recognition with insufficient data, but this is beyond the scope of this paper.

\textbf{Language Modeling:} Pre-training language models is a foundational problem for NLP. ELMo~\cite{peters2018deep} was proposed to learn deep contextualized word representations. It was trained with a bidirectional language model objective, which combined both a forward and backward language model. ELMo representations significantly improve the performance across six NLP tasks. GPT~\cite{radford2018improving} employed a standard language model objective to pre-train a language model on large unlabeled text corpora. The Transformer was used as the model architecture. The pre-trained model was fine-tuned on downstream tasks and achieved excellent results in 9 of 12 tasks. BERT~\cite{devlin2018bert} used a masked language model pre-training objective, which enabled BERT to learn bidirectional representations conditioned on the left and right context.  BERT employed a multi-layer bidirectional Transformer encoder and advanced the state-of-the-art performance. 
Our work is inspired by the recent success of these methods and is the first attempt to model multi-attribute learning from the perspective of language modeling.

\textbf{Transformer for Computer Vision:} Transformer~\cite{vaswani2017attention} was first proposed for sequence modeling in NLP. Recently, Transformer-based methods have been deployed in many computer vision tasks~\cite{bao2021beit,kaiming2021masked,yu2021frequency,liu2021swin,wang2021end,zhang2021temporal,perrett2021temporal}. ViT~\cite{dosovitskiy2020image} demonstrated that a pure transformer architecture achieved very competitive results on image classification tasks. DETR~\cite{carion2020end} formulated the object detection as a set prediction problem and employed a transformer encoder-decoder architecture. Pix2Seq~\cite{chen2021pix2seq} regarded object detection as a language modeling task and obtained competitive results. Zheng \emph{et al.}~\cite{zheng2021rethinking} replaced the encoder of FCN with a pure transformer
for semantic segmentation. Liu \emph{et al.}~\cite{liu2021query2label} utilized the Transformer decoder architecture for multi-label classification. Temporal query networks were introduced in~\cite{zhang2021temporal} for fine-grained video understanding with a query-response mechanism. 
There are also some efforts~\cite{lanchantin2021general,cheng2021mltr,nguyen2021modular} to apply Transformer to the task of multi-label image classification.
Note that the main contribution of this paper is not the use of Transformer, but modeling multi-attribute recognition from the perspective of language modeling.

\section{Approach}
In this section, we first give an overview of our framework. Then we present the details of the proposed attribute query network and image-conditioned masked language model. Lastly, we introduce the training objective function and inference process of our method.

\begin{figure}[t]
  \begin{center}
     \includegraphics[width=1.0\linewidth]{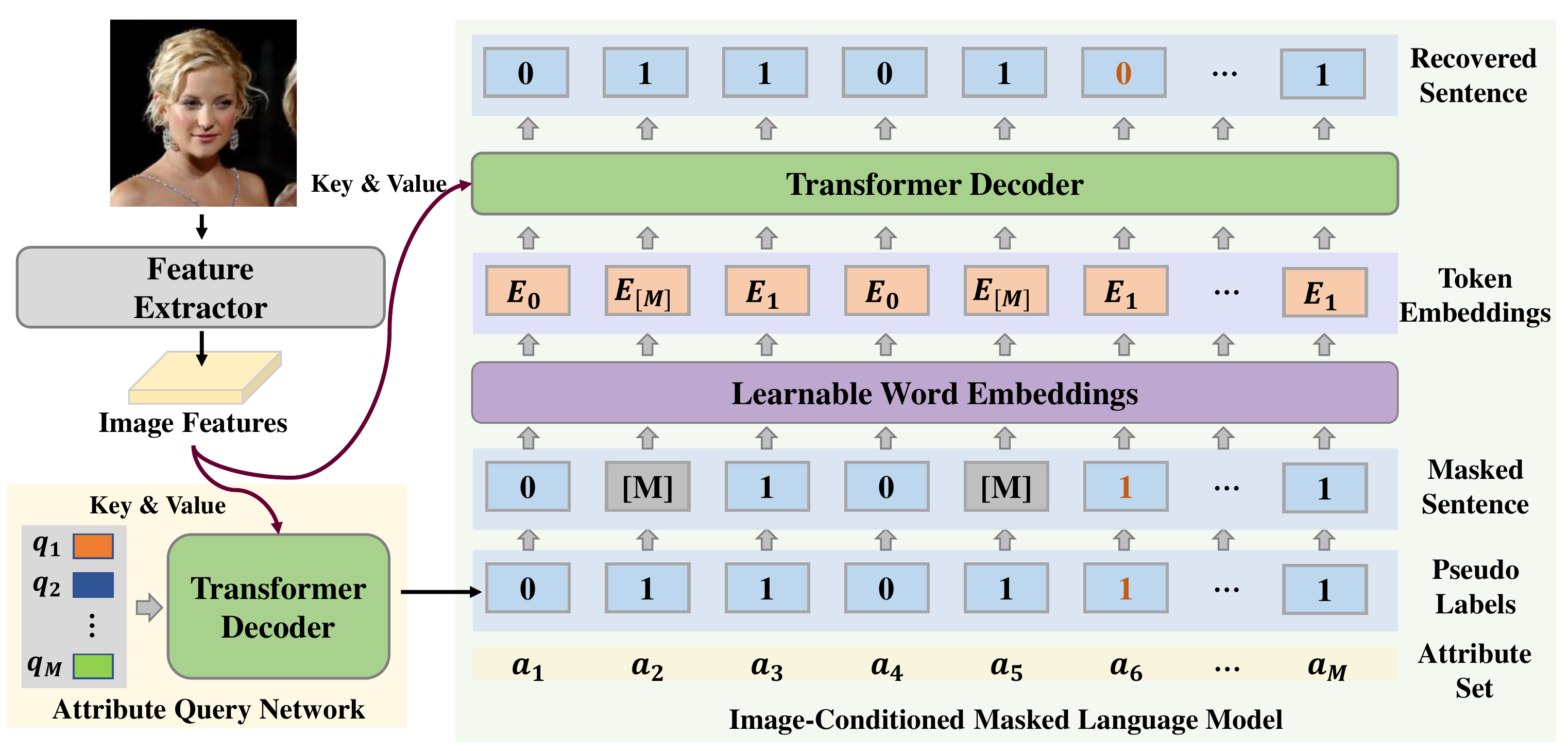}
  \end{center}
     \caption{The pipeline of our framework. 
     We recover the entire label ``sentence'' with a Transformer decoder module, which is conditioned on the token embeddings and image features. Although there are some wrong ``words'' in the pseudo labels, which are shown in orange, we can treat them as another form of masks. Here $\bm{E_1}$ or $\bm{E_0}$ indicates the presence or absence of an attribute.}
  \label{fig:flowchart}
  \end{figure}

\subsection{Overview}
Given a sample $\bm{x}$ from a dataset $\mathcal{D}$ with $M$ attribute types, we aim to predict the multiple attributes $\bm{y}$ to the image $\bm{x}$. We let $\mathcal{A} = \{\bm{a}_1,\bm{a}_2, ..., \bm{a}_M\}$ denote the attribute set, where $\bm{a}_j (1 \le j \le M)$ represents the $j$-th attribute type. For simplicity, we assume that the values of all attribute types are binary. In other words, the value of $\bm{a}_j$ is $0$ or $1$, where $1$ means that the sample has this attribute and $0$ means not. However, our method can be easily extended to the case where each attribute type is multi-valued. With this assumption, we have $\bm{y} \in \{0,1\}^M$. Existing methods~\cite{jia2021spatial,cao2018partially} usually employ a multi-tasking learning framework, which uses $M$ binary classifiers to predict $M$ attributes respectively. Binary cross-entropy loss is used as the objective.

This paper proposes a language modeling framework. We show the pipeline of our framework in Fig. \ref{fig:flowchart}. The key idea of this paper is to treat attribute labels as unordered ``sentences'' and use an image-conditioned masked language model to exploit the relationships between attributes. Although we can directly use the real attribute labels as the input of the IC-MLM during training, we cannot access these labels for inference. To address this issue, our Label2Label introduces an attribute query network to generate the initial attribute predictions. These predictions are then treated as pseudo-labels and used as input to the IC-MLM in the training and testing phases.

\subsection{Attribute Query Network}

Given an input image $\bm{x} \in \mathbb{R}^{H_0 \times W_0 \times 3}$ and its corresponding label $\bm{y} =\{y_j| 1\le j \le M\}$, we send the image to a feature extractor to obtain the image features, where $H_0$ and $W_0$ denote the height and width of the input image respectively, $y_{j}$ denotes the value of $j$-th attribute $\bm{a}_j$ for the sample $\bm{x}$. As our framework is agnostic to the feature extractor, we can use any popular backbones such as  ResNet-50~\cite{he2016deep} and ViT~\cite{dosovitskiy2020image}.  A naive way to generate initial attribute predictions is to directly feed the extracted image features to a linear layer and learn $M$ binary classifiers. As recent progress~\cite{zhang2021temporal,doersch2020crosstransformers,lanchantin2021general,liu2021query2label} shows the superiority of Transformer, we consider using the Transformer decoder to implement our attribute query network to generate initial predictions with higher quality.

Our attribute query network learns a set of permutation-invariant  query vectors $\bm{Q} = \{\bm{q}_1, \bm{q}_2, ..., \bm{q}_M \}$, where each query $\bm{q}_j$ corresponds to an attribute type $\bm{a}_j$. Then each query vector $\bm{q}_j$ pools the attribute-related features from the image features with Transformer decoder layers and generates the corresponding response vector $\bm{r}_j$. Finally, we learn a binary classifier for each response vector to generate the initial attribute predictions.

Since many attributes are only located in some local areas of the image, using global image features is not an excellent choice. Therefore, we preserve the spatial dimensions of image features following \cite{liu2021query2label}. For ResNet-50, we simply abandon the global pooling layer and employ the output of the last convolution block as the extracted features. We denote the extracted features as $\bm{X} \in \mathbb{R}^{H \times W \times d}$, where $H$, $W$, and $d$ represent the height, width, and channel of the image features respectively. To fit with the Transformer decoder, we reshape the feature to be $\bm{X}' \in \mathbb{R}^{HW \times d}$. Following common practices~\cite{carion2020end,dosovitskiy2020image}, we add 2D-aware position embeddings $\bm{X}_{pos} \in \mathbb{R}^{HW \times d}$ to the feature vectors $\bm{X}'$ to retain positional information. In this way, we obtain the visual feature vectors $\bm{\widetilde{X}} = \bm{X}' + \bm{X}_{pos}$.

With the local visual contexts $\bm{\widetilde{X}}$, the query features $\bm{Q} = \{\bm{q}_j \in \mathbb{R}^{d} |1 \le j \le M\}$ are updated using multi-layer Transformer decoders. Formally, we update the query features $\bm{Q}_{i-1}$ in the $i$-th Transformer decoder layer as follows:
\begin{equation}
  \begin{split}
    &\bm{Q}_{i-1}^{sa} = \mathrm{MultiHead}(\bm{Q}_{i-1},\bm{Q}_{i-1},\bm{Q}_{i-1}),\\
    &\bm{Q}_{i-1}^{ca} =  \mathrm{MultiHead}(\bm{Q}_{i-1}^{sa},\bm{\widetilde{X}},\bm{X}'),\\
    &\bm{Q}_{i} =  \mathrm{FFN}(\bm{Q}_{i-1}^{ca}),\\
  \end{split}
  \label{equ:attdecoder}
\end{equation}
where the $\mathrm{MultiHead}()$ and $\mathrm{FFN}()$ denote the multi-head attention layer and feed-forward layer respectively. Here we set $\bm{Q}$ as $\bm{Q}_0$.
The design philosophy is that for each attribute query vector, it can give high attention scores to the interested local visual features to produce attribute-related features. This design is compatible with the locality of some attributes. Assuming that the attribute query network consists of $L$ layers of Transformer decoders, then we denote $\bm{Q}_{L}$ as $\bm{R} = \{\bm{r}_{1}, \bm{r}_{2}, ..., \bm{r}_{M}\}$, where each response vector $\bm{r}_{j} \in \mathbb{R}^{d}$ corresponds to a query vector $\bm{q}_j$.
With the response vectors, we use $M$ independent binary classifiers to predict the attribute values $l_{j} = \sigma(\bm{W}_j^T \bm{r}_{j} + b_{j})$,
where $\bm{W}_j \in \mathbb{R}^d$ and $b_j \in \mathbb{R}^1$ are learnable parameters of the $j$-th attribute classifier, $\sigma(\cdot)$ is the sigmoid function and $l_{j}$ is the predicted probability for attribute $\bm{a}_j$ of image $\bm{x}$.
In the end, we read out the pseudo label ``sentence'' $\bm{s} = \{s_{1},\bm{s}_{2}, ..., \bm{s}_{M} \}$ from the predictions $\{l_{j}\}$ with $\bm{s}_{j} = \mathbb I (l_{j} > 0.5)$,
where $\mathbb I(\cdot)$ is an indicator function.

It is worth noting that the predictions from the attribute query network are not $100\%$ correct, resulting in some wrong ``words'' in the generated label ``sentence''. However, we can treat the wrong ``words'' as another form of masks, because the wrong predictions account for only a small proportion. In fact, the masking strategy of the wrong word is artificially performed in some language models, such as BERT~\cite{devlin2018bert}.

\subsection{Image-Conditioned Masked Language Model}
In existing multi-attribute databases, images are annotated with a variety of attribute labels. This paper is dedicated to modeling sample-wise complex attribute correlations.
Instead of treating attribute labels as numbers, we regard them as ``words''. Since different attribute labels describe the object in an image from different perspectives, we can group them as a sequence of ``words''. Although the sequence is essentially an unordered ``sentence'' without any grammatical structure, it still conveys meaningful contextual information. In this way, we treat $\bm{y}$ as an unordered yet meaningful ``sentence'', where $y_{j}$ is a ``word''.

By treating the labels as sentences, we resort to language modeling methods to mine the instance-level attribute relations effectively. In recent years, pre-training large-scale task-agnostic language models have substantially advanced the development of NLP, among which representative works include ELMo~\cite{peters2018deep}, GPT-3~\cite{brown2020language}, BERT~\cite{devlin2018bert}, and so on. Inspired by the success of these methods, we consider a masked language model to learn the relationship between ``words''. We mask some percentage of the attribute label ``sentence'' $\bm{y}$ at random, and then reconstruct the entire label ``sentence''.  Specifically, for a binary label sequence, we replace those masked ``words'' with a special work token \texttt{[mask]} to obtain the masked sentence. Then we input the masked sentence to a masked language model, which aims to recover the entire label sequence. While the MLM has proven to be an effective tool in NLP, directly using it for multi-attribute learning is not feasible. Therefore, we propose several important improvements.

\textbf{Instance-wise Attribute Relations:} MLM essentially constructs the task $P(y_1,y_2,..., y_M|\mathcal{M}(y_1),\mathcal{M}(y_2),...,\mathcal{M}(y_M))$ to capture the ``word'' co-occurrence and learn the joint probability of ``word'' sequences $P(y_1,y_2,..., y_M)$, where $\mathcal{M}()$ denotes the random masking operation. Such a naive approach leads to two problems. The first problem is that MLM only captures statistical attribute correlations. A diverse dataset means that the mapping $\{ \mathcal{M}(y_1),\mathcal{M}(y_2),...,\mathcal{M}(y_M)\} \mapsto \{y_1,y_2,..., y_M\}$ is a one-to-many mapping. Therefore MLM only learns how different attributes are statistically related to each other. Meanwhile, our experiments find that this prior can be easily modeled by the attribute query network $P(y_1,y_2,..., y_M |\bm{x})$. The second problem is that MLM and attribute query network cannot be jointly trained. Since MLM uses only the hard prediction of the attribute query network, the gradient from MLM cannot influence the training of the attribute query network. In this way, the method becomes a two-stage label refinement process, which significantly reduces the optimization efficiency.

To address these issues, we propose an image-conditioned masked language model to learn instance-wise attribute relations. Our IC-MLM captures the relations by constructing a task $P(y_1,y_2,..., y_M|\bm{x}, \mathcal{M}(y_1),\mathcal{M}(y_2),...,\mathcal{M}(y_M))$. Introducing an extra image condition is not trivial, as this fundamentally changes the behavior of MLM. With the conditions of image $\bm{x}$, the transformation $\{ \bm{x},\mathcal{M}(y_1),\mathcal{M}(y_2),...,\mathcal{M}(y_M)\} \mapsto \{y_1,y_2,..., y_M\}$ is an accurate one-to-one mapping. Our IC-MLM infers other attribute values by combining some attribute label prompts and image contexts in the precise image-to-label mapping, which facilitates the model to learn sample-level attribute relations. In addition, IC-MLM and the attribute query network can use shared image features, which enables them to be jointly optimized with a one-stage framework.

\textbf{Word Embeddings:}
It is known that the word id is not a good word representation in NLP. Therefore, we need to map the word id to a token embedding. Instead of utilizing existing word embeddings with a large token vocabulary like BERT~\cite{devlin2018bert}, we directly learn attribute-related word embeddings $\bm{E}$ from
scratch. We use the word embedding module to map the ``word'' in the masked sentence to the corresponding token embedding. Since all attributes are binary, we need to build a token vocabulary with a size of $2M$ to model all possible attribute words. Also, we need to include the token embedding for the special word \texttt{[mask]}. This paper considers three different strategies for the \texttt{[mask]} token embedding. The first strategy believes the \texttt{[mask]} words for different attributes have different meanings, so $M$ attribute-specific learnable token embeddings are learned, where one \texttt{[mask]} token embedding corresponds to one attribute. The second strategy treats the \texttt{[mask]} words for different attributes as the same word. Only one attribute-agnostic learnable token embedding is learned and shared by all attributes. The third strategy is based on the second strategy, which simply replaces the learnable token embedding with a fixed $\bm{0}$ vector. Our experiments find all three strategies work well while the first strategy performs best.

As mentioned earlier, we use pseudo labels $\bm{s} = \{s_{1},\bm{s}_{2}, ..., \bm{s}_{M} \}$ as input to IC-MLM, so we actually construct $P(y_1,y_2,..., y_M|\bm{x},\mathcal{M}(\bm{s}_1),\mathcal{M}(\bm{s}_2),...,\mathcal{M}(\bm{s}_M))$ as the task. We randomly mask out some ``words'' in the pseudo-label sequence with a probability of $\alpha$ to generate masked label ``sentences''. The ``word'' $\mathcal{M}(\bm{s}_j)$ in the masked label ``sentences'' may have three values: 0, 1, and \texttt{[mask]}.
We use the word embedding module to map the masked labels ``sentences'' to a sequence of token embeddings $\bm{E} = \{\bm{E}_{1},\bm{E}_{2}, ..., \bm{E}_{M}  \}$ according to the word value,
where $\bm{E}_{j} \in \mathbb{R}^d$ denotes the embedding for ``word'' $\mathcal{M}(\bm{s}_j)$.

\textbf{Positional Embeddings:} In BERT, the positional embedding of each word is added to its corresponding token embeddings to obtain the position information.
Since our ``sentences'' are unordered, there is no need to introduce positional embeddings to ``word'' representations. We conducted experiments with positional embeddings by randomly defining some word order and found no improvement. Therefore we do not use positional embeddings for ``word'' representations and the learned model is permutation invariant for ``words''.

\textbf{Architecture:} In NLP, Transformer encoder layers are usually used to implement MLM, while we use multi-layer Transformer decoders to implement IC-MLM due to additional image input conditions. Following the design philosophy similar to the attribute query network, token embeddings $\bm{E}$ pool features from the local visual features $\bm{X}'$ with a cross-attention mechanism. We update the token features $\bm{E}_{i-1}$ in the $i$-th Transformer decoder layer as follows:
\begin{equation}
  \begin{split}
    &\bm{E}_{i-1}^{sa} = \mathrm{MultiHead}(\bm{E}_{i-1},\bm{E}_{i-1},\bm{E}_{i-1}),\\
    &\bm{E}_{i-1}^{ca} =  \mathrm{MultiHead}(\bm{E}_{i-1}^{sa},\bm{\widetilde{X}},\bm{X}'),\\
    &\bm{E}_{i} =  \mathrm{FFN}(\bm{E}_{i-1}^{ca}).\\
  \end{split}
  \label{equ:labeldecoder}
\end{equation}

We set $\bm{E}$ to $\bm{E}_0$ and the number of Transformer decoder layers in IC-MLM to $D$.
Then we denote $\bm{E}_D$ as $\bm{R}' = \{\bm{r}'_{1}, \bm{r}'_{2}, ..., \bm{r}'_{M}\}$, where $\bm{r}'_{j}$ corresponds to the updated feature of token $\bm{E}_j$.
In the end, we perform the final multi-attribute classification with linear projection layers.
Formally, we have:
\begin{equation}
p_{j} = \sigma({\bm{W}_j'}^T \bm{r}'_{j} + b'_{j}),  1 \le j \le M,
\label{equ:labelcls}
\end{equation}
where $\bm{W}'_j \in \mathbb{R}^d$ and $b'_j \in \mathbb{R}^1$ are the learnable parameters of the $j$-th attribute classifier, and $p_{j}$ is the final predicted probability for attribute $\bm{a}_j$ of image $\bm{x}$.
Note that we are committed to recovering the entire label ``sentence'' and not just the masked part. In this reconstruction process, we expect our model to grasp the instance-level attribute relations.

\subsection{Objective and Inference}
As commonly used in most existing methods~\cite{jia2021spatial,sarafianos2018deep,liu2015deep}, we adopt the binary cross-entropy loss to train the IC-MLM. On the other hand, since most of the datasets for multi-attribute recognition are highly imbalanced, different tasks usually use different weighting strategies. The loss function for the IC-MLM is formulated as $\mathcal{L}_{mlm}(\bm{x}) \!\!=\!\!\sum_{j=1}^{M} \!w_{j}(y_{j} \!\log(p_{j}) \!+\! (1 \!-\! y_{j}) \!\log(1\!-\!p_{j} ))$,
where $w_{j}$ is the weighting coefficient. According to different tasks, we choose different weighting strategies and always follow the most commonly used strategy for a fair comparison. Meanwhile, to ensure the quality of the generated pseudo label sequences, we also supervise the attribute query network with the same loss function $\mathcal{L}_{aqn}(\bm{x}) \!\!=\!\!\sum_{j=1}^{M} \!w_{j}(y_{j} \!\log(l_{j}) \!+\! (1 \!-\! y_{j}) \!\log(1\!-\!l_{j} ))$.
The final loss function $\mathcal{L}_{total}$ is a combination of the two loss functions above:
\begin{equation}
\mathcal{L}_{total}(\bm{x}) = \mathcal{L}_{aqn}(\bm{x}) + \lambda \mathcal{L}_{mlm}(\bm{x}),
\label{equ:allloss}
\end{equation}
where $\lambda$ is used to balance these two losses.
At inference time, we ignore the masking step and directly input the pseudo label ``sentence'' to the IC-MLM. Then the output of the IC-MLM is used as the final attribute prediction.


\section{Experiments}
In this section, we conducted extensive experiments on three multi-attribute learning tasks to validate the effectiveness of the proposed framework.

  \begin{table*}[tb]
    \begin{minipage}[t]{0.475\textwidth}
    \centering\small
        \caption{ Results with different Transformer decoder layers $D$ for IC-MLM. We fix $L$ as 1.}
        \label{table:ablation:Llbl}
          \renewcommand\tabcolsep{3pt}
          \begin{tabular}{lcccc}
            \toprule
            $D$ &  1 &  2 & 3 & 4 \\
            \midrule
            Error(\%) &  12.58 & \textbf{12.49} & 12.54 & 12.52 \\
            \bottomrule
            \end{tabular}
    \end{minipage}
  \hfill
    \begin{minipage}[t]{0.475\textwidth}
    \centering\small
        \caption{Results with different Transformer decoder layers $L$ for attribute query network. We fix $D$ as 2.}
        \label{table:ablation:Latr}
          \renewcommand\tabcolsep{3pt}
          \begin{tabular}{lcccc}
            \toprule
            $L$ &  1 &  2 & 3 & 4 \\
            \midrule
            Error(\%) & \textbf{12.49} & 12.52 & 12.50 & 12.58 \\
            \bottomrule
            \end{tabular}
    \end{minipage}
  \end{table*}

\begin{table*}[tb]
  \begin{minipage}[t]{0.475\textwidth}
  \centering\small
      \caption{
        Results on the LFWA dataset with different mask ratios $\alpha$. 
      }
      \label{table:ablation:alpha}
        \renewcommand\tabcolsep{1pt}
        \begin{tabular}{lccccc}
          \toprule
          $\alpha$ & 0 &  0.1 & 0.15 & 0.2 &  0.3 \\
          \midrule
           Error(\%) & 12.55 & \textbf{12.49} & 12.55 & 12.54 & 12.57 \\
          \bottomrule
          \end{tabular}
  \end{minipage}
\hfill
  \begin{minipage}[t]{0.475\textwidth}
  \centering
      \caption{
        Results on the LFWA dataset with different coefficients $\lambda$.
      }
      \label{table:ablation:lambda}
        \renewcommand\tabcolsep{1pt}
        \begin{tabular}{lccccc}
          \toprule
          $\lambda$ & 0.5 &  0.8 & 1 & 1.2 &  1.5 \\
          \midrule
           Error(\%) & 12.64 & 12.56 & \textbf{12.49} & 12.60 & 12.63 \\
          \bottomrule
          \end{tabular}
  \end{minipage}
\end{table*}

\subsection{Facial Attribute Recognition}
\textbf{Dataset:} LFWA~\cite{liu2015deep}
is a popular unconstrained facial attribute dataset, which
consists of 13,143 facial images of 5,749 identities. Each facial image has 40 attribute annotations. Following the same evaluation protocol in~\cite{liu2015deep,cao2018partially,hand2017attributes}, we partition the LFWA dataset into two sets, with 6,263 images for training and 6,880 for testing. All images are pre-cropped to a size of $250 \times 250$. We adopt the classification error for evaluation following~\cite{cao2018partially,shu2021learning}. 

\textbf{Experimental Settings:} We trained our model for 57 epochs with a batch size of 16. For optimization, we used an SGD optimizer with a base learning rate of 0.01 and cosine learning rate decay. The weight decay was set to 0.001. To augment the dataset, Rand-Augment~\cite{Cubuk2020RandAugment} and Random horizontal flipping were performed. We also adopted Mixup~\cite{zhang2017mixup} for regularization.

\begin{table}[t]
  \caption{Ablation experiments with different backbones.}
  \label{table:ablation:back}
  \centering
  \begin{tabular}{lcccccc}
  \toprule
  Backbone & \multicolumn{2}{c}{ResNet-50}& \multicolumn{2}{c}{ResNet-101}  & \multicolumn{2}{c}{ViT-B} \\
  \midrule
  Metric & Error(\%) & MACs(G) & Error(\%) & MACs(G) & Error(\%) & MACs(G) \\ 
  \midrule
  FC Head & 13.63$\pm$0.02 & 5.30 & 13.05$\pm$0.03 & 10.15 & 13.73$\pm$ 0.02& 16.85  \\
  AQN & 13.36$\pm$0.04 & 5.63 & 12.70$\pm$0.02 & 10.48 & 13.32$\pm$0.04 & 16.97 \\
  Label2Label & \textbf{12.49}$\pm$0.02& 6.30 & \textbf{12.44}$\pm$0.04 & 11.16& \textbf{12.79}$\pm$0.01 & 17.23\\
  \bottomrule
  \end{tabular}
\end{table}

\textbf{Parameters Analysis:} We first analyze the influence of the number of Transformer decoder layers in the attribute query network and IC-MLM. The results are shown in Tables \ref{table:ablation:Llbl} and \ref{table:ablation:Latr}. We see that the best performance is achieved when $L =1$ and $D = 2$. We further conduct experiments with different mask ratios $\alpha$ and list the results in Table \ref{table:ablation:alpha}. As we mentioned above, the wrong ``words'' in the pseudo label sequences also provide some form of masks. Therefore, our method performs well when $\alpha = 0$. We observe that our method attains the best performance when $\alpha =0.1$. Table \ref{table:ablation:lambda} shows the results with different $\lambda$, and we see that $\lambda = 1$ gives the best trade-off in \eqref{equ:allloss}. We consider three different strategies for \texttt{[MASK]} token embedding and list the results in Table \ref{table:ablation:mask}. We see that the 
attribute-specific strategy achieves the best performance among them, as it better models the differences between the attributes.
Unless explicitly mentioned, we adopt these optimal parameters in all subsequent experiments.

  \begin{table*}[t]
    \begin{minipage}[t]{0.375\textwidth}
    \centering\small
        \caption{
         Results of different strategies for \texttt{[Mask]} embeddings.
        }
        \label{table:ablation:mask}
          \renewcommand\tabcolsep{1pt}
          \begin{tabular}{lc}
            \toprule
            Strategy &  Error(\%)   \\
            \midrule
            $\bm{0}$ Vector & 12.60\\
            Attribute-Agnostic &12.57\\
            Attribute-Specific &\textbf{12.49}\\
            \bottomrule
            \end{tabular}
    \end{minipage}
  \hfill
    \begin{minipage}[t]{0.575\textwidth}
    \centering
        \caption{
          Comparisons of MLM and IC-MLM.
        }
        \label{table:ablation:mlm}
          \renewcommand\tabcolsep{1pt}
          \begin{tabular}{lccc}
            \toprule
            \multirow{2}*{Method} & \multirow{2}*{Architecture} & Co-training  & \multirow{2}*{Error(\%)} \\
            ~ & ~ &  with AQN & ~ \\
            \midrule
            \multirow{2}*{MLM} &  MLP & \XSolidBrush & 13.34  \\
            ~ &  TransEncoder &\XSolidBrush & 13.32 \\
            \midrule
            \multirow{2}*{IC-MLM} & TransDecoder &\XSolidBrush  & 13.01 \\
            ~ &  TransDecoder & \Checkmark & \textbf{12.49} \\
            \bottomrule
            \end{tabular}
    \end{minipage}
  \end{table*}

\textbf{Ablation Study:} 
To validate the effectiveness of our Label2Label, we also conduct experiments on the LFWA dataset with two baseline methods. We first consider the Attribute Query Network (AQN) method, which ignores the IC-MLM and treats the outputs of AQN in Fig.~\ref{fig:flowchart} as final predictions. FC Head method further replaces the Transformer decoder layers in AQN with a linear classification layer. To further verify the generalization of our method, we use different feature extraction backbone networks for ablation experiments. To better demonstrate the significance of the results, we also report the standard deviation. The results are presented in Table \ref{table:ablation:back}.  In addition, we report the computation cost (MACs) of each method in Table \ref{table:ablation:back}.
We observe that our method significantly outperforms FC Head and AQN across various backbones with marginal computational overhead, which illustrates the effectiveness of our method.

\setlength{\intextsep}{0pt}
\begin{wraptable}{r}{.5\linewidth}
	\centering 
	\caption{Performance comparison with state-of-the-art methods on the LFWA dataset. We report the average classification error results. * indicates that additional labels are used for training, such as identity labels or segment annotations.}
	\label{table:result:lfw}
	\begin{tabular}{lcc}
    \toprule
    Method & Error(\%) &  Year\\
    \midrule
    SSP + SSG~\cite{kalayeh2017improving}* & 12.87 & 2017 \\
    He \emph{et al.}~\cite{he2018harnessing} & 14.72 & 2018 \\
    AFFAIR~\cite{li2018landmark} & 13.87 & 2018 \\
    GNAS~\cite{huang2018gnas} & 13.63 & 2018 \\
    PS-MCNN~\cite{cao2018partially}* & 12.64 & 2018 \\
    DMM-CNN~\cite{mao2020deep} & 13.44 & 2020\\
    SSPL~\cite{shu2021learning} & 13.47 & 2021 \\
    \midrule
    Label2Label & \textbf{12.49}$\pm$0.02 & - \\
    \bottomrule
    \end{tabular}
\end{wraptable}

We then conducted experiments to show how image-conditioned MLM improves performance. The results are listed in Table \ref{table:ablation:mlm}. As we analyzed above, MLM leads to a two-stage label refinement process. We consider two network architectures to implement MLM: Transformer encoder and multilayer perceptron (MLP). The results show that none of them improve the performance of AQN (13.36\%). The reason is that MLM only learns statistical attribute relations, and this prior is easily captured by AQN. Meanwhile, our IC-MLM learns instance-wise attribute relations. To see the benefits of the additional image conditions, we still adopt the two-stage label refinement process, and train Transformer decoder layers with fixed image features. 
We see that performance is boosted to 13.01\%, which demonstrates the effectiveness of modeling instance-wise attribute relations. We further jointly train the IC-MLM and attribute query network, which achieves significant performance improvement. These results illustrate the superiority of the proposed IC-MLM.

\textbf{Comparison with State-of-the-art Methods:} Following~\cite{shu2021learning}, we employ ResNet50 as the backbone. We present the performance comparison on the LFWA dataset in Table \ref{table:result:lfw}. We observe that our method attains the best performance with a simple framework compared to highly tailored domain-specific methods. Label2Label even exceeds the methods~\cite{kalayeh2017improving,cao2018partially} of using additional annotations, which further illustrates the effectiveness of our framework.

\begin{figure}[t]
\begin{center}
   \includegraphics[width=1.0\linewidth]{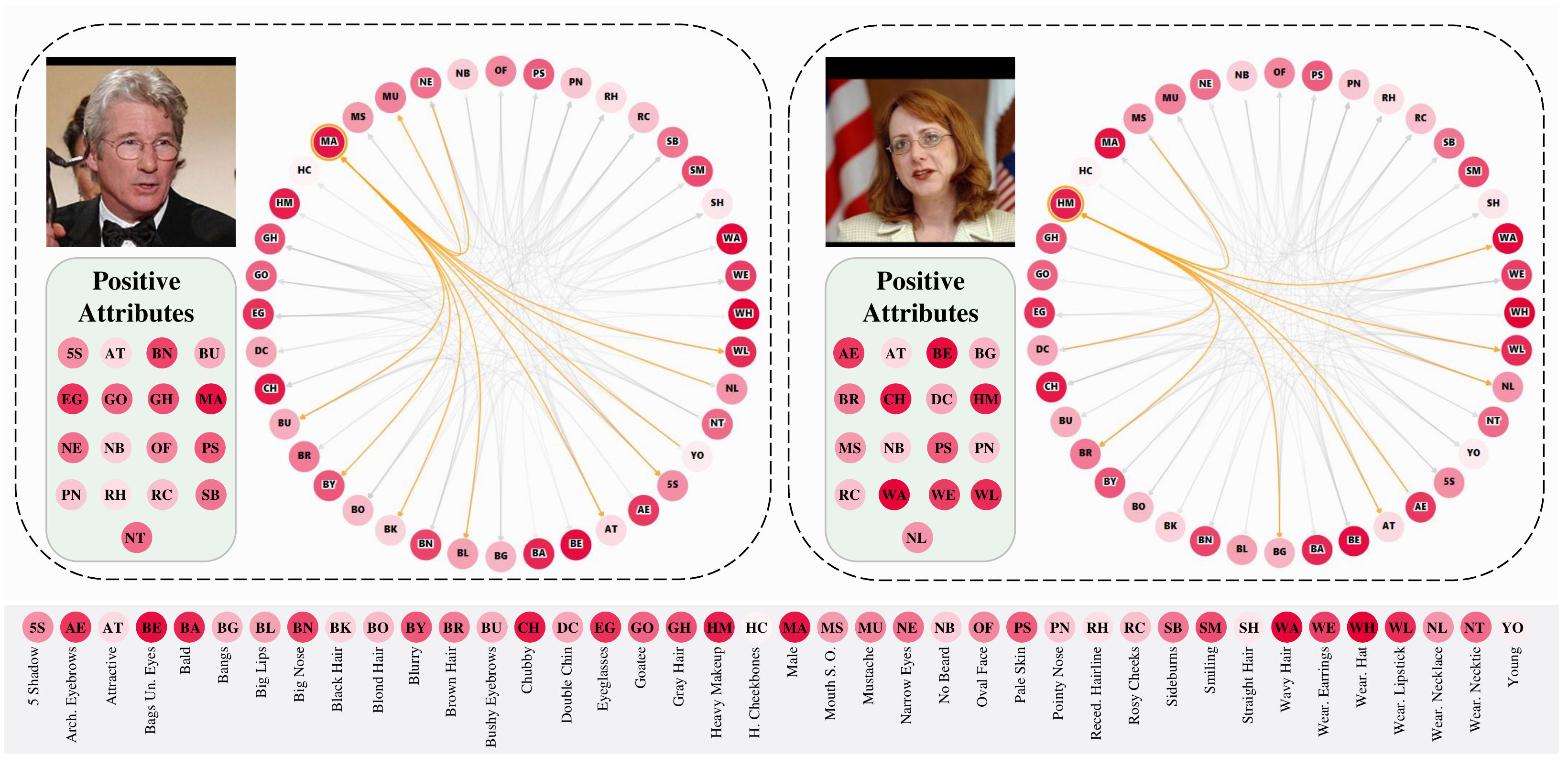}
\end{center}
   \caption{Visualization of attention scores among attributes in the self-attention layer. We show the attention of the first head at layer 1 with two samples. The positive attributes of each sample are listed in the corresponding bottom-left corner. 
}
\label{fig:visual}
\end{figure}

\textbf{Visualization:} As the Transformer decoder architecture is used to model the instance-level relations, our method can give better interpretable predictions. We visualize the attention scores in the IC-MLM with DODRIO~\cite{wang2021dodrio}. As shown in Fig.~\ref{fig:visual}, we see that related attributes tend to have higher attention scores.

\subsection{Pedestrian Attribute Prediction}

\begin{table}[t]
  \caption{
    Comparisons on the PA100K dataset. * represents the reimplementation performance using the same setting. We also report the standard deviations.
  }
\label{table:result:pa100k}
\renewcommand\tabcolsep{4pt}
\centering
\begin{tabular}{lcccccc}
\toprule
Method  & mA & Accuracy & Precision & Recall & F1 \\
\midrule
DeepMAR~\cite{li2015multi} & 72.70 & 70.39 & 82.24 & 80.42 & 81.32 \\
HPNet~\cite{liu2017hydraplus} & 74.21 & 72.19 & 82.97 & 82.09 & 82.53 \\
VeSPA~\cite{sarfraz2017deep} & 76.32& 73.00& 84.99& 81.49& 83.20 \\
LGNet~\cite{liu2018localization} &76.96 &75.55& 86.99& 83.17& 85.04 \\
PGDM~\cite{li2018pose} & 74.95 & 73.08 & 84.36 & 82.24  & 83.29 \\
\midrule
MsVAA~\cite{sarafianos2018deep}* & 80.10 & 76.98 & 86.26 & 85.62 & 85.50 \\
VAC~\cite{guo2019visual}* & 79.04 & 78.95 & \textbf{88.41} & 86.07 & 86.83 \\
ALM~\cite{tang2019improving}* &  79.26  & 78.64  & 87.33 & 86.73 & 86.64 \\
SSC~\cite{jia2021spatial} & 81.87 & 78.89 & 85.98  & \textbf{89.10} & 86.87 \\
\midrule
FC Head & 77.96$\pm$0.06 & 75.86$\pm$0.79 & 86.27$\pm$0.13 & 84.16$\pm$1.02 & 84.72$\pm$0.55 \\
AQN & 80.89$\pm$0.08 & 78.51$\pm$0.08 & 86.15$\pm$0.40 & 87.85$\pm$0.43 & 86.58$\pm$0.03 \\
Label2Label & \textbf{82.24}$\pm$0.13 & \textbf{79.23}$\pm$0.13 & 86.39$\pm$0.32 & 88.57$\pm$0.20 & \textbf{87.08}$\pm$0.08 \\
\bottomrule
\end{tabular}
\end{table}

\textbf{Dataset:} The PA-100K~\cite{liu2017hydraplus} dataset is
the largest pedestrian attribute dataset so far~\cite{tang2019improving}. It contains 100,000 pedestrian images from 598 scenes, which are collected from real outdoor surveillance videos. All pedestrians in each image are annotated with 26 attributes including gender, handbag, and upper clothing. The dataset is randomly split into three subsets: 80\% for training, 10\% for validation, and 10\% for testing. Following SSC~\cite{jia2021spatial}, we merge the training set and the validation set for model training.
We use five metrics: one label-based and four instance-based. For the label-based metric, we adopt the mean accuracy (mA) metric.
For instance-based metrics, we employ accuracy, precision, recall, and F1 score. As mentioned in~\cite{tang2019improving}, mA and F1 score are more appropriate and convincing criteria for class-imbalanced pedestrian attribute datasets.

\textbf{Experimental Settings:} Following the  state-of-the-art methods~\cite{jia2021spatial,guo2019visual}, we adopted ResNet50 as the backbone network to extract image features. We first resize all images into 256$\times$192 pixels. Then random flipping and random cropping were used for data augmentation. SGD optimizer was utilized with the weight decay of 0.0005. We set the initial learning rate of the backbone to 0.01. For fast convergence, we set the initial learning rate of the attribute query network and IC-MLM to 0.1. The batch size was equal to 64. We trained our model for 25 epochs using a plateau learning rate scheduler. We reduced the learning rate by a factor of 10 once learning stagnates and the patience was 4.

\textbf{Results and Analysis:} We report the results in  Table~\ref{table:result:pa100k}. We observe that Label2Label achieves the best performance in mA, Accuracy, and F1 score. Compared to the previous state-of-the-art method SSC~\cite{jia2021spatial}, which designs complex SPAC and SEMC modules to extract discriminative semantic features, our method achieves 0.37\% performance improvements in mA. In addition, we report the re-implemented results of the MsVAA, VAC, and ALM methods in the same setting as did in~\cite{jia2021spatial}. Our method consistently outperforms these methods.
We further show the results of the FC Head and Attribute Query Network. 
We see that the performance is improved by replacing the FC head with Transformer decoder layers, 
which shows the superiority of our attribute query network.
Our Label2Label outperforms the attribute query network method by 1.35\% for mA, which shows the effectiveness of the language modeling framework.

\subsection{Clothing Attribute Recognition}
\textbf{Dataset:} Clothing Attributes Dataset~\cite{chen2012describing} consists of 1,856 images that contain clothed people. Each image is annotated with 26 clothing attributes, such as colors and patterns. We use 1,500 images for training and the rest for testing. For a fair comparison, we only use 23 binary attributes and ignore the remaining three multi-class value attributes as in~\cite{abdulnabi2015multi,meng2018efficient}. We adopt 
accuracy as the metric and also report the accuracy of four clothing attribute groups following~\cite{abdulnabi2015multi,meng2018efficient}.

\begin{table}[tbp]
  \caption{The comparisons between our method and other state-of-the-art methods on the Clothing Attributes Dataset. 
  We report accuracy and standard deviation.}
\label{table:result:clothing}
\centering
\begin{tabular}{lccccc}
\toprule
Method  & Colors & Patterns & Parts & Appearance & Total \\
\midrule
S-CNN~\cite{abdulnabi2015multi} & 90.50 &92.90& 87.00& 89.57& 90.43 \\
M-CNN~\cite{abdulnabi2015multi} & 91.72& 94.26& 87.96& 91.51& 91.70 \\
MG-CNN~\cite{abdulnabi2015multi} & 93.12& 95.37& 88.65& 91.93& 92.82 \\
Meng \emph{et al.}~\cite{meng2018efficient} & 91.64& 96.81& 89.25& 89.53& 92.39 \\
\midrule
FC Head & 91.39$\pm$0.23&  96.07$\pm$0.05&  87.00$\pm$0.27&  88.21$\pm$0.36 & 91.57$\pm$0.12 \\
AQN &  91.98$\pm$0.25& 96.37$\pm$0.23& 88.19$\pm$0.47&  89.89$\pm$0.33&  92.29$\pm$0.05 \\
Label2Label & 92.73$\pm$0.07 & 96.82$\pm$0.02& 88.20$\pm$0.09& 90.88$\pm$0.18& \textbf{92.87}$\pm$\textbf{0.03}\\
\bottomrule
\end{tabular}
\end{table}

\textbf{Experimental Settings:} For a fair comparison, we utilized AlexNet to extract image features following~\cite{abdulnabi2015multi,meng2018efficient}. We trained our model for 22 epochs using a cosine decay learning rate scheduler. We utilized an SGD optimizer with an initial learning rate of 0.05.
The batch size was set to 32. For the attribute query network, we employed a 2-layer Transformer decoder ($L=2$).

\textbf{Results and Analysis:} 
Table \ref{table:result:clothing} shows the results.
We observe that our Label2Label attains a total accuracy of 92.87\%, which outperforms other methods with a simple framework. MG-CNN learns one CNN for each attribute, resulting in more training parameters and longer training time. Compared with the attribute query network method, our method achieves better performance on all attribute groups, which illustrates the superiority of our framework.

\section{Conclusions}
In this paper, we have presented Label2Label, which is a simple and generic framework for multi-attribute learning. Different from the existing multi-task learning framework, we proposed a language modeling framework, which regards each attribute label as a ``word''. 
Our model learns instance-level attribute relations by the proposed image-conditioned masked language model, which randomly masks some ``words'' and restores them based on the remaining ``sentence'' and image context. Compared to well-optimized domain-specific methods, Label2Label attains competitive results on three multi-attribute learning tasks.

\noindent \textbf{Acknowledgments.} 
This work was supported in part by the National Key Research and Development Program of China under Grant 2017YFA0700802, in part by the National Natural Science Foundation of China under Grant 62125603 and Grant U1813218, in part by a grant from the Beijing Academy of Artificial Intelligence (BAAI).  The authors would sincerely thank Yongming Rao and Zhiheng Li for their generous helps.

\newpage
\appendix
\begin{center}
\noindent{\textbf{\large{Supplementary Materials}}}
\end{center}

\section{Evaluation Metrics}
For pedestrian attribute prediction, we adopted five evaluation metrics. We present the details of these metrics. The only label-based metric is the mean accuracy (mA) metric, which is the mean of positive accuracy and negative accuracy for each attribute. Mathematically, the mA is calculated by:
\begin{equation}
mA = \frac{1}{2M} \sum_{j=1}^{M} (\frac{TP_j}{P_j} + \frac{TN_j}{N_j}),
\label{equ:metric:mA}
\end{equation}
where $M$ is the number of attributes, $P_j$ and $TP_j$ represent the numbers of positive samples and correctly predicted positive samples of the $j$-th attribute respectively, $N_j$ and $TN_j$ are the numbers of negative samples and correctly predicted negative samples of the $j$-th attribute respectively.

We also consider four example-based metrics: accuracy, precision, recall, and F1 score:
\begin{equation}
\label{equ:metric:instance}
  \begin{split}
  Acc&= \frac{1}{N} \sum_{i=1}^{N} \frac{|\bm{Y}_i \cap \bm{Y}'_i|}{|\bm{Y}_i  \cup \bm{Y}'_i|},  Prec = \frac{1}{N} \sum_{i=1}^{N} \frac{|\bm{Y}_i \cap \bm{Y}'_i|}{ | \bm{Y}'_i|},   \\
  Rec&= \frac{1}{N} \sum_{i=1}^{N} \frac{|\bm{Y}_i \cap \bm{Y}'_i|}{|\bm{Y}_i  |}, F1 = \frac{2 * Prec * Rec}{Prec + Rec},
  \end{split}
\end{equation}
where $N$ denotes the number of samples, $\bm{Y}_i$ is the  positive labels of the $i$-th sample and $\bm{Y}'_i$ is the predicted positive values for the $i$-th sample.

\begin{figure}[t]
\begin{center}
   \includegraphics[width=1.0\linewidth]{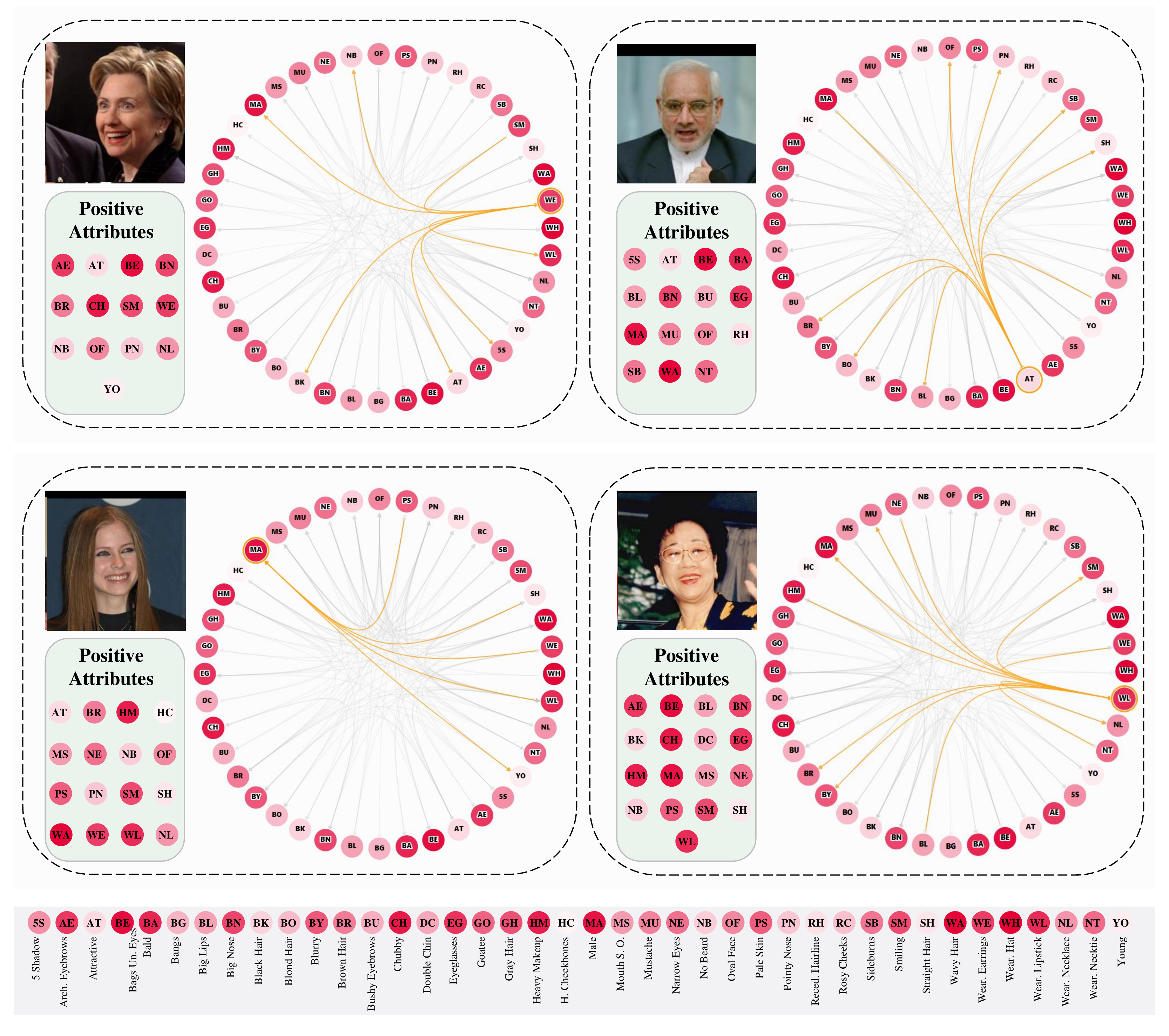}
\end{center}
\caption{More visualization results of attention scores in the self-attention layer. We show the attention of the first head at layer 1 with four samples. The positive ground truth attribute labels of each sample are listed in the corresponding bottom-left corner.
}
\label{fig:morevisual}
\end{figure}

\section{Weighting Strategy}
For facial attribute recognition and clothing attribute recognition, we follow the common practice which does not utilize the weighting strategy for loss functions. Therefore, we have:
\begin{equation}
\label{equ:loss:noweight}
  \begin{split}
  \mathcal{L}_{mlm}(\bm{x})&=\sum_{j=1}^{M} \!y_{j} \!\log(p_{j}) \!+\! (1 \!-\! y_{j}) \!\log(1\!-\!p_{j} ),  \\
   \mathcal{L}_{aqn}(\bm{x}) &=\sum_{j=1}^{M} \!y_{j} \!\log(l_{j}) \!+\! (1 \!-\! y_{j}) \!\log(1\!-\!l_{j} ).
  \end{split}
\end{equation}

For pedestrian attribute recognition, we follow the widely used
weighted binary-entropy strategy in~\cite{jia2021spatial,tang2019improving}. In this way, we have:
\begin{equation}
\label{equ:loss:weight}
  \begin{split}
  \mathcal{L}_{mlm}(\bm{x}) \!\!&=\!\!\sum_{j=1}^{M} \!w_{j}(y_{j} \!\log(p_{j}) \!+\! (1 \!-\! y_{j}) \!\log(1\!-\!p_{j} )),  \\
   \mathcal{L}_{aqn}(\bm{x}) \!\!&=\!\!\sum_{j=1}^{M} \!w_{j}(y_{j} \!\log(l_{j}) \!+\! (1 \!-\! y_{j}) \!\log(1\!-\!l_{j} )), \\
   w_{j} &= y_{j} e^{1- \gamma_j} + (1 -  y_{j} )e^{\gamma_j},
  \end{split}
\end{equation}
where $\gamma_j$ is the positive example ratio of the $j$-th attribute.

\section{More Ablation Studies}
\begin{table}[t]
    \begin{minipage}[t]{0.375\linewidth}
        \caption{
          \small{Ablation experiments on the position embeddings of word representations.}
        }
        \label{table:ablation:pos_word_embed}
        \begin{center}
          \renewcommand\tabcolsep{4pt}
          \begin{tabular}{lcc}
            \toprule
            Method & Pos &  Error(\%)   \\
            \midrule
            \multirow{2}*{Label2Label} & \XSolidBrush & 12.49 \\
            ~ & \Checkmark & 12.51 \\
            \bottomrule
            \end{tabular}
        \end{center}
    \end{minipage}
  \hfill
    \begin{minipage}[t]{0.575\linewidth}
        \caption{
          \small{Ablation experiments on the position embeddings of visual features. }
        }
        \label{table:ablation:pos_visual_embed}
        \begin{center}
          \renewcommand\tabcolsep{6pt}
          \begin{tabular}{lcc}
            \toprule
            Method & Pos &  Error(\%) \\
            \midrule
            \multirow{2}*{AQN} & \XSolidBrush & 13.51  \\
            ~ &\Checkmark & \textbf{13.36} \\
            \midrule
            \multirow{2}*{Label2Label} &\XSolidBrush  & 12.98 \\
            ~ & \Checkmark & \textbf{12.49} \\
            \bottomrule
            \end{tabular}
        \end{center}
    \end{minipage}
  \end{table}
  
\subsection{Position Embeddings for Word Representations}
We conducted ablation experiments on the position embeddings of word representations. Since we are dealing with unordered ‘‘sentences’’, we randomly define three different label sequences and use the corresponding position embeddings respectively. We report the average performance of  three different label sequences on the LFWA database in Table \ref{table:ablation:pos_word_embed}. We found no additional performance gain from the position embeddings of the word representations. The reason is that our ‘‘sentences’’ are essentially made up of unordered ‘‘words’’.

\subsection{Position Embeddings for Visual Features}
In our paper, we add 2D-aware position embeddings to visual feature vectors to retain positional information. We conduct experiments to verify their effectiveness and show the results on the LFWA database in Table \ref{table:ablation:pos_visual_embed}. We observe that introducing position embeddings in visual features is beneficial for performance.

\subsection{Comparisons with Transformer-based Multi-label Classification Methods}
Many Transformer-based multi-label classification methods~\cite{lanchantin2021general,liu2021query2label} have been proposed in recent years. To further verify the effectiveness of the proposed method, we conducted experiments on the three datasets used in our paper. Table \ref{table:comparison1} shows the results. We see our method consistently outperforms C-Tran~\cite{lanchantin2021general} and Q2L~\cite{liu2021query2label}, which shows the superiority of our method.

\begin{table}[t]
\caption{Comparisons of our method with other Transformer-based methods. }
\begin{center}
\renewcommand\tabcolsep{10pt}
\begin{tabular}{l|c|c|c|c|c}
\hline\hline
Dataset  & LFWA  &\multicolumn{3}{c|}{PA100K} & Clothing\\
\hline
Meteic & Error & mA & Accuracy & F1 & Accuracy\\
\hline 
C-Tran \cite{lanchantin2021general} & 14.66 &  81.53 & 78.97  &  86.86 &  90.00  \\
Q2L \cite{liu2021query2label} & 13.28 &  80.72  &  78.78 &  86.73 &  91.81 \\
\hline
Ours & \textbf{12.49} & \textbf{82.37} &  \textbf{79.03} & \textbf{86.96} &  \textbf{92.87} \\
\hline\hline
\end{tabular}
\end{center}
\label{table:comparison1}
\end{table}

\subsection{The Need of Masking}
To verify the effectiveness of masking, we construct three pure reconstruction (without masking) baselines. 1) Feature Reconstruction: direct reconstruction of the word features $\bm{r}_1, \bm{r}_2,...,\bm{r}_M$. 2) Score Reconstruction: direct reconstruction of the predicted scores $l_1,l_2,...,l_M$. 3) Label Reconstruction:  direct reconstruction of the labels: $y_1,y_2,...y_M$. Table \ref{table:masking} shows the results on the LFWA dataset. Although the Label Reconstruction works competitively, it is still inferior to our method with masking.  Just as found in ~\cite{he2022masked}, although Autoencoder (reconstruction) works well, the Masked Autoencoder (masking) is the key factor to learning better features.
In BERT, the masked word is replaced with the \texttt{[mask]} token or a random word. So the MLM has two tasks: mask-recovering and error-correcting.
Both increase the training difficulty. In our method,
the wrong predictions are like random words in BERT. See Table \ref{table:masking}, Ours ($\alpha=0$) outperforms Label Reconstruction (12.55 vs 12.70). The only difference is that the input of our IC-MLM contains wrong predictions while Label Reconstruction does not, which proves that our proposed IC-MLM also benefits from handling this special ``mask''.

\begin{table}[t]
\caption{Comparisons with three pure reconstruction baselines. }
\begin{center}
\renewcommand\tabcolsep{10pt}
\begin{tabular}{l|c|c|c|c|c}
\hline\hline
\multirow{2}*{Method} &  \multicolumn{3}{c|}{Reconstruction} & \multicolumn{2}{c}{Ours (Masking)}\\ 
\cline{2-6}
& Feature & Score & Label &$\alpha=0$ &$\alpha=0.1$ \\
\hline 
Error(\%) & 13.45 & 13.63 & 12.70  &  12.55 &  \textbf{12.49}  \\
\hline\hline
\end{tabular}
\end{center}
\label{table:masking}
\end{table}

\section{Network Structure Configuration}
We show the default hyper-parameters for the Transformer decoder layer of our method in Table \ref{Hyperparameters}.

\begin{table}[htbp]
    \centering
    \renewcommand\tabcolsep{10pt}
    \vspace{5pt}
    \caption{Hyperparameters for the Transformer decoder layer of our method.}
    \label{Hyperparameters}
    \begin{tabular}{cc}
        \toprule[1pt]
        Component & Hyperparameters \\
        \midrule 
        Activation & GELU\\
        Hidden dim & 2048 \\
        FFN hidden size & 2048\\
        Attention heads & 4\\
        Attention head szie & 512 \\
        \bottomrule[1pt]
    \end{tabular}
\end{table}

\section{More Visualization Results}
We provide more visualization results of the attention scores in Figure~\ref{fig:morevisual}. We conducted the experiments on the LFWA database. We read out the attention from the self-attention layer of our label decoder. The DODRIO~\cite{wang2021dodrio} is used for visualization. We show the attention scores of the first head at layer 1 with four examples.

For the first example, the attribute ``Wearing Earrings'' is strongly related to the existence of ``Wearing Lipstick'', ``No Beard'', and ``Female'' and the absence of ``5 o'clock shadow''. For the second sample, the attributes ``Oval Face'', ``Pointy Nose'', ``Sideburns'', ``Wearing Necktie'' and ``Male'' imply the existence of ``Attractive''. For the third sample, the attributes ``Wearing Earrings'' and ``Wearing Lipstick'' indicate the gender ``Female''. For the last example, the attribute ``Wearing Lipstick'' assigns more attention to the existence of ``Wearing Earrings'', ``Wearing Necklace'', ``Heavy Makeup'' and the absence of ``Mustache'', ``Male''.  We see our method can learn the instance-level attribute relations even if a sample has some wrong labels.

\begin{table}[thbp]
	\setlength{\abovecaptionskip}{0.cm}
	\setlength{\belowcaptionskip}{-0.cm}
	\centering
	\caption{ The classification error (\%) obtained by all the competing methods on the LFWA datasets. The accuracy for each attribute obtained by the proposed method is highlighted in bold.}
	\scriptsize
	\setlength{\tabcolsep}{0.1mm}{
	\begin{tabular}{c|c|c|c|c|c|c|c|c|c|c|c|c|c|c}
		\toprule 
		& \rotatebox{90}{5 o'clock Shadow}	&	\rotatebox{90}{Arched Eyebrows}	&	\rotatebox{90}{Attractive}	&	\rotatebox{90}{Bags Under Eyes}	&	\rotatebox{90}{Bald}	&	\rotatebox{90}{Bangs}	&
		\rotatebox{90}{Big Lips}	&	\rotatebox{90}{Big Nose}	&	\rotatebox{90}{Black Hair}	&	\rotatebox{90}{Blond Hair}	&	\rotatebox{90}{Blurry}	&	\rotatebox{90}{Brown Hair}	&	\rotatebox{90}{Bushy Eyebrows}	&	\rotatebox{90}{Chubby}\\
		\midrule
		  PANDA~\cite{zhang2014panda} & 16.00 & 21.00 & 19.00 & 20.00 & 16.00 & 16.00 & 27.00 & 21.00 & 13.00 & 6.00 & 26.00 & 26.00 & 21.00 & 31.00 \\

		 LNets+ANet~\cite{liu2015deep}	&	16.00 & 18.00 & 17.00 & 17.00 & 12.00 & 12.00 & 25.00 & 19.00 & 10.00 & 3.00 & 26.00 & 23.00 & 18.00 & 27.00  \\												
		 NSA~\cite{mahbub2018segment} & 22.41 & 18.28 & 19.84 & 17.38 & 8.12 & 9.29 & 21.03 & 16.87 & 7.51 & 2.53 & 13.58 & 19.07 & 15.74 & 23.94  \\																			
		 MCNN-AUX~\cite{hand2017attributes}	&	22.94 & 18.22 & 19.69 & 16.52 & 8.06 & 9.92 & 20.76 & 15.02 & 7.37 & 2.59 & 14.77 & 19.15 & 15.03 & 23.14 \\		
     MCFA~\cite{zhuang2018multi} & 25.00 & 21.00 & 23.00 & 21.00 & 9.00 & 11.00 & 25.00 & 19.00 & 9.00 & 3.00 & 14.00 & 23.00 & 24.00 & 26.00 \\	
 		
     {PS-MCNN-LC~\cite{cao2018partially}} & 21.83 & 16.47 & 18.16 & 13.26 & 7.40 & 8.55 & 17.30 & 13.52 & 7.04 & 1.49 & 12.80 & 18.13 & 14.28 & 21.89 \\		
 
     DMTL~\cite{han2017heterogeneous} & 20.00 & 14.00 & 18.00 & 16.00 & 8.00 & 7.00 & 23.00 & 17.00 & 8.00 & 3.00 & 11.00 & 19.00 & 20.00 & 25.00 	\\																
    DMM-CNN~\cite{mao2020deep}	&	20.82 & 17.30 & 18.90 & 17.30 & 8.04 & 8.70 & 20.18 & 16.33 & 8.45 & 2.83 & 12.42 & 18.44 & 14.67 & 22.34  \\																			
    Label2Label &  \textbf{20.76} & \textbf{16.67} & \textbf{18.28} & \textbf{16.10} & \textbf{6.93} & \textbf{8.06} & \textbf{19.40} & \textbf{15.00} & \textbf{6.96} & \textbf{2.25} & \textbf{13.00} & \textbf{16.88}& \textbf{12.89} & \textbf{21.61} \\
		\midrule

		&	\rotatebox{90}{Double Chin}	&	\rotatebox{90}{Eyeglasses}	&	\rotatebox{90}{Goatee}	&	\rotatebox{90}{GrayHair}	&	\rotatebox{90}{Heavy Makeup}	&	\rotatebox{90}{High Cheekbones}	&	\rotatebox{90}{Male}   &	\rotatebox{90}{MouthOpen}	&	\rotatebox{90}{Mustache}	&	\rotatebox{90}{NarrowEyes}	&	\rotatebox{90}{NoBeard}	&	\rotatebox{90}{OvalFace}	&	\rotatebox{90}{PaleSkin}	&	\rotatebox{90}{PointyNose} \\
		\midrule 
	PANDA~\cite{zhang2014panda} & 25.00 & 11.00 & 25.00 & 19.00 & 7.00 & 14.00 & 8.00 &	22.00 & 13.00 & 27.00 & 25.00 & 28.00 & 16.00 & 24.00 \\
	LNets+ANet~\cite{liu2015deep} & 22.00 & 5.00 & 22.00 & 16.00 & 5.00 & 12.00 & 6.00 & 18.00 & 8.00 & 19.00 & 21.00 & 26.00 & 16.00 & 20.00 \\
	NSA~\cite{mahbub2018segment} & 19.51 & 8.50 & 16.99 & 11.54 & 4.61 & 11.66 & 7.40 & 17.50 & 7.03 & 17.25 & 19.23 & 23.20 & 9.03 & 15.80  \\
	MCNN-AUX~\cite{hand2017attributes}  & 18.48 & 8.70 & 17.03 & 11.07 & 4.15 & 11.62 & 5.98 & 6.49 & 6.57 & 17.14 & 17.85 & 22.61 & 6.68 & 15.86 \\
	MCFA~\cite{zhuang2018multi} & 23.00 & 9.00 & 20.00 & 12.00 & 6.00 & 15.00 & 7.00 & 22.00 & 9.00 & 22.00 & 21.00 & 26.00 & 18.00 & 20.00 \\
	{PS-MCNN-LC~\cite{cao2018partially}} & 13.30 & 7.22 & 15.89 & 8.96 & 3.40 & 11.23 & 4.82 & 15.40 & 5.53 & 16.49 & 17.99 & 22.10 & 5.03 & 12.48 \\
	DMTL~\cite{han2017heterogeneous} & 22.00 & 8.00 & 14.00 & 12.00 & 5.00 & 11.00 & 7.00 & 14.00 & 5.00 & 18.00 & 19.00 & 25.00 & 9.00 & 16.00 \\
	DMM-CNN~\cite{mao2020deep} & 19.02 & 7.17 & 17.18 & 10.62 & 4.32 & 11.87 & 5.86 & 15.55 & 5.54 & 16.33 & 17.52 & 23.06 & 8.14 & 15.49  \\
	Label2Label  & \textbf{16.24} & \textbf{7.38} & \textbf{15.34} & \textbf{10.13} & \textbf{3.88} & \textbf{10.36} & \textbf{5.78} & \textbf{16.38} & \textbf{5.89} & \textbf{15.52} & \textbf{16.45} & \textbf{20.26} & \textbf{8.47} & \textbf{15.09} \\
	\midrule
	
	  	&	\rotatebox{90}{RecedingHairline}	&	\rotatebox{90}{RosyCheeks}	&	\rotatebox{90}{Sideburns}	&	\rotatebox{90}{Smiling}	&	\rotatebox{90}{Straight Hair}	&	\rotatebox{90}{WavyHair}	&	\rotatebox{90}{WearingEarrings}	&	\rotatebox{90}{WearingHat}	&	\rotatebox{90}{WearingLipstick}	&	\rotatebox{90}{WearingNecklace}	&	\rotatebox{90}{WearingNecktie}	&	\rotatebox{90}{Young}	&		&	\rotatebox{90}{Average} \\
	    \midrule
	
	    PANDA~\cite{zhang2014panda}  & 16.00 & 27.00 & 24.00 & 11.00 & 27.00 & 25.00 & 8.00 & 18.00 & 7.00 & 14.00 & 21.00 & 18.00 &   & 18.97
	    \\
	    LNets+ANet~\cite{liu2015deep} & 15.00 & 22.00 & 23.00 & 9.00 & 24.00 & 24.00 & 6.00 & 12.00 & 5.00 & 12.00 & 21.00 & 14.00 &   & 16.15	\\
        
        NSA~\cite{mahbub2018segment}  & 15.10 & 12.92 & 18.24 & 9.20 & 21.09 & 21.72 & 5.25 & 9.77 & 5.93 & 10.41 & 18.60 & 14.32 &   & 14.18\\	  
        
        MCNN-AUX~\cite{hand2017attributes}	 & 13.75 & 12.08 & 16.87 & 8.17 & 21.47 & 18.39 & 5.05 & 9.93 & 4.96 & 10.06 & 19.34 & 14.16 &   & 13.69	\\	
	    
       MCFA~\cite{zhuang2018multi}  & 15.00 & 15.00 & 22.00 & 12.00 & 23.00 & 21.00 & 7.00 & 9.00 & 6.00 & 11.00 & 18.00 & 13.00 &   & 16.37	\\		
       {PS-MCNN-LC~\cite{cao2018partially}} & 12.50 & 11.19 & 15.58 & 7.30 & 20.35 & 16.65 & 4.46 & 8.79 & 4.30 & 9.08 & 17.82 & 13.12 &   & 12.64\\	
       
	    DMTL~\cite{han2017heterogeneous}  & 15.00 & 14.00 & 20.00 & 8.00 & 21.00 & 20.00 & 6.00 & 8.00 & 7.00 & 9.00 & 19.00 & 13.00 &   & 13.85 	\\											
	    DMM-CNN~\cite{mao2020deep}	& 13.70 & 13.56 & 17.01 & 7.76 & 20.80 & 20.13 & 5.86 & 9.16 & 4.89 & 10.53 & 18.72 & 11.06 &   & 13.44	\\	
	    
        Label2Label  & \textbf{12.46} & \textbf{10.65} & \textbf{14.90} & \textbf{7.85} & \textbf{16.83} & \textbf{17.17} & \textbf{5.12} & \textbf{8.02} & \textbf{4.96} & \textbf{9.74} & \textbf{16.03} & \textbf{13.82}& & \textbf{12.49} \\	
		\bottomrule
	\end{tabular}}
	\label{tab:comprehensive}
\end{table}

\section{Detailed Results} 

For facial attribute recognition, some methods~\cite{mao2020deep,han2017heterogeneous} report the pre-class recognition accuracy. We report the pre-attribute classification error on the LFWA database in Table~\ref{tab:comprehensive} for a comprehensive comparison.

We observe our method attains very competitive results with a simple framework compared to highly tailored domain-specific methods, which demonstrates the effectiveness of our method.

%
%
\bibliographystyle{splncs04}
\bibliography{egbib}

\end{document}